\documentclass[conference,a4paper]{APSIPA2012}

\usepackage{graphicx}
\usepackage{amssymb,amsmath,bm}
\usepackage{textcomp}
\usepackage{epsfig}
\usepackage{epstopdf, color}

\title{Multi-task Recurrent Model for True Multilingual Speech Recognition}

\author{%
\authorblockN{%
Zhiyuan Tang\authorrefmark{2}\authorrefmark{3},
Lantian Li\authorrefmark{2} and
Dong Wang\authorrefmark{2}\authorrefmark{1}
}
\authorblockA{%
\authorrefmark{2}
Center for Speech and Language Technologies, Division of Technical Innovation and Development, \\
Tsinghua National Laboratory for Information Science and Technology\\
Center for Speech and Language Technologies, Research Institute of Information Technology, Tsinghua University }
\authorblockA{%
\authorrefmark{3}
Chengdu Institute of Computer Applications, Chinese Academy of Sciences\\
E-mail: \{tangzy, lilt\}@cslt.riit.tsinghua.edu.cn\\
$^*$Corresponding author: wangdong99@mails.tsinghua.edu.cn}%
}

\begin{document}

  \maketitle
  \begin{abstract}

  Research on multilingual speech recognition remains attractive yet challenging.
  Recent studies focus on learning shared structures under the multi-task paradigm,
  in particular a feature sharing structure.
  This approach has been found effective to improve performance on each \emph{individual} language.
  However, this approach is only useful when the deployed system supports
  just one language. In a true multilingual scenario where multiple languages are
  allowed, performance will be significantly reduced due to the competition among
  languages in the decoding space.

  This paper presents a multi-task recurrent model
  that involves a multilingual speech recognition (ASR) component and a language recognition (LR)
  component, and the ASR component is informed of the language information by the
  LR component, leading to a language-aware recognition.
  We tested the approach on an English-Chinese bilingual recognition task.
  The results show that the proposed multi-task recurrent model can improve
  performance of multilingual recognition systems.

  \end{abstract}

  \section{Introduction}

  Speech recognition (ASR) technologies develop fast in recent years,
  partly due to the powerful deep learning approach~\cite{hinton2012deep, yu2015automatic}.
  An interesting and important task within the ASR research
  is recognizing multiple languages. One reason that makes
  the multilingual ASR research attractive is that people
  from different countries are communicating
  more frequently today. Another reason is that
  there are limited resources for most languages,
  and multilingual techniques may help to improve performance
  for these low-resource languages.

  There has been much work on multilingual ASR,
  especially with the deep neural architecture.
  The mostly studied architecture is the feature-shared
  deep neural network (DNN), where the input and
  low-level hidden layers are shared across languages,
  while the top-level layers and the output layer are
  separated for each language~\cite{huang2013cross, Ghoshal2013multilingual, Heigold2013multilingual}.
  The insight of this design is that the human languages
  share some commonality in both acoustic and phonetic
  layers, and so some signal patterns
  at some levels of abstraction can be shared.

  Despite the brilliant success of the feature-sharing approach,
  it is only useful for model training, not for decoding.
  This means that although part of the model structure is shared,
  in recognition (decoding), the models are used independently
  for individual languages, with their own language models.
  Whenever more than one language are supported, the performance
  on all the languages will be significantly decreased, due to the
  inter-language competition in the decoding process. This means that
  the feature-sharing approach cannot deal with true multilingual ASR,
  or more precisely, multilingual decoding.


  A possible solution to the multilingual decoding problem is to
  inform the decoder which language it is now processing. By this
  language information, the multilingual decoding essentially falls back
  to monolingual decoding and the performance is recovered. However,
  language recognition is subject to recognition mistakes, and it requires
  sufficient signal to give a reasonable inference, leading to unacceptable
  delay. Another possibility is to invoke monolingual decoding for each
  language, and then decide which result is correct, due to either confidence or
  a language recognizer. This approach obviously requires more computing resource.
  In Deepspeech2~\cite{amodei2015deep}, English and Chinese can be jointly decoded under
  the end-to-end learning framework. However, this is based on the fact that
  the training data for the two languages are both abundant, so that language
  identities can be learned by the deep structure. This certainly can not
  be migrated to other low-resource languages, and is difficult to accommodate
  more languages.

  In this paper, we introduce a multi-task recurrent model for multilingual decoding.
  With this model, the ASR model and the LR model are treated as two components of
  a unified architecture, where the output of one component is propagated back
  to the other as extra information. More specifically,
  the ASR component provides speech information for the LR
  component to deliver more accurate language information, which in turn
  helps the ASR component to produce better results. Note that
  this collaboration among ASR and LR takes places in both model training and
  inference (decoding).

  This model is particularly attractive for multilingual decoding.
  By this model, the LR component provides language information for the ASR
  component when decoding an utterance.
  This language information is produced frame by frame, and becomes more
  and more accurate when the decoding proceeds. With this information, the
  decoder becomes more and more confident about which language it is processing, and
  gradually removes decoding paths in hopeless languages.
  Note that the multi-task recurrent model was proposed in~\cite{tang2016multi},
  where we found that it can learn speech and speaker recognition models in a
  collaborative way. The same idea was also proposed by~\cite{li2015modeling}, though it
  focused on ASR only. This paper tests the approach
  on an English-Chinese bilingual recognition task.


  The rest of the paper is organized as follows: Section~\ref{sec:model} describes
  the model architectures, and Section~\ref{sec:exp} reports the experiments.
  The conclusions plus the future work are presented in Section~\ref{sec:con}.

  \section{Models}
  \label{sec:model}

  Consider the feature-sharing bilingual ASR.
  Let $x$ represent the primary input feature, $t_1$ and $t_2$ represent
  the targets for each language respectively, and $c$ is the extra input
  obtained from other component (LR in our experiments).
  With the information $c$,
  the model estimates the probability $P(t_1|x, c)$ and $P(t_2|x, c)$ respectively, that
  makes the decoding of two languages absolutely separate. $P(t|x, c)$ is truly required
  by  multilingual decoding, where $t$ means the targets for both two languages.
  If we regard the extra input $c$ as a language indicator,
  the model is language-aware.
  Note that the language-aware model is a conditional model with the context $c$ as the condition.
  In contrast, the feature-sharing model, which can be formulated
  as $P(t_1|x)$ or $P(t_2|x)$, is essentially a marginal model $\sum_{c}P(t_2|x,c)P(c|x)$ or $\sum_{c}P(t_2|x,c)P(c|x)$,
  which are more complex and less effective for listing $c$.

  We refer the bilingual ASR as a single task,
  with respect to the single task of LR.
  So $P(t|x, c)$ is what we actually compute with the proposed model jointly training ASR and LR, that indicates
  the two languages use the same Gaussian Mixture Model (GMM) system for generative modeling, though
  the two languages still use their own phone sets.

  We first describe the
  single-task baseline model and
  then multi-task recurrent model as in~\cite{tang2016multi}.

  \subsection{Basic single-task model}

\begin{figure}[ht]
        \centering
        \includegraphics[width=1\linewidth]{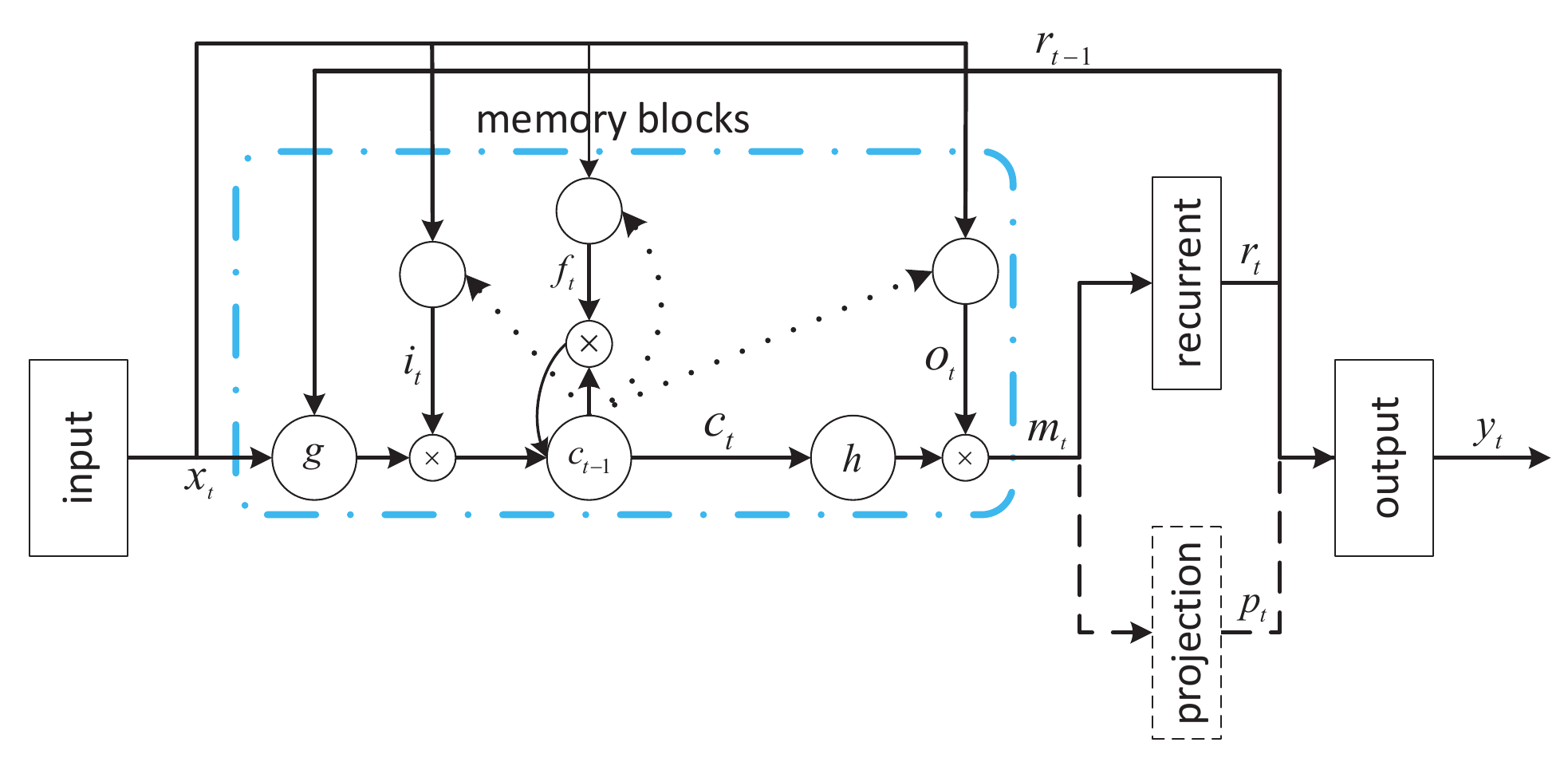}
        \caption{ Basic recurrent LSTM model for ASR and LR single-task baselines.
        The picture is reproduced from~\cite{Sak2014long1}.}
        \label{fig:lstm}
  \end{figure}

  The most promising architecture
  for ASR is the recurrent neural network, especially the long short-term memory (LSTM)~\cite{Sak2014long1, Sak2014long2}
  for its ability of modeling temporal sequences and their long-range dependencies.
  The modified LSTM structure proposed in~\cite{Sak2014long1} is used. The network structure is shown in Fig.~\ref{fig:lstm}.

  The associated computation is as follows:
  \begin{eqnarray}
    i_t &=& \sigma(W_{ix}x_{t} + W_{ir}r_{t-1} + W_{ic}c_{t-1} + b_i) \nonumber\\
    f_t &=& \sigma(W_{fx}x_{t} + W_{fr}r_{t-1} + W_{fc}c_{t-1} + b_f) \nonumber\\
    c_t &=& f_t \odot c_{t-1} + i_t \odot g(W_{cx}x_t + W_{cr}r_{t-1} + b_c) \nonumber\\
    o_t &=& \sigma(W_{ox}x_t + W_{or}r_{t-1} + W_{oc}c_t + b_o) \nonumber\\
    m_t &=& o_t \odot h(c_t) \nonumber\\
    r_t &=& W_{rm} m_t \nonumber\\
    p_t &=& W_{pm} m_t \nonumber\\
    y_t &=& W_{yr}r_t + W_{yp}p_t + b_y \nonumber
  \end{eqnarray}

  \noindent In the above equations, the $W$ terms denote weight matrices and those associated with cells were set to be diagonal in our
  implementation. The $b$ terms denote bias
  vectors. $x_t$ and $y_t$ are the input and output symbols respectively; $i_t$, $f_t$, $o_t$ represent respectively
  the input, forget and output gates; $c_t$ is the cell and $m_t$ is the cell output. $r_t$ and $p_t$ are two output components derived from $m_t$, where $r_t$ is recurrent and fed to the next time step, while $p_t$ is not recurrent and contributes to the present output only.
  $\sigma(\cdot)$ is the logistic sigmoid function, and $g(\cdot)$ and $h(\cdot)$ are non-linear activation functions, often chosen to be hyperbolic. $\odot$ denotes the element-wise multiplication.

  \subsection{Multi-task recurrent model}

\begin{figure}[ht]
        \centering
        \includegraphics[width=1\linewidth]{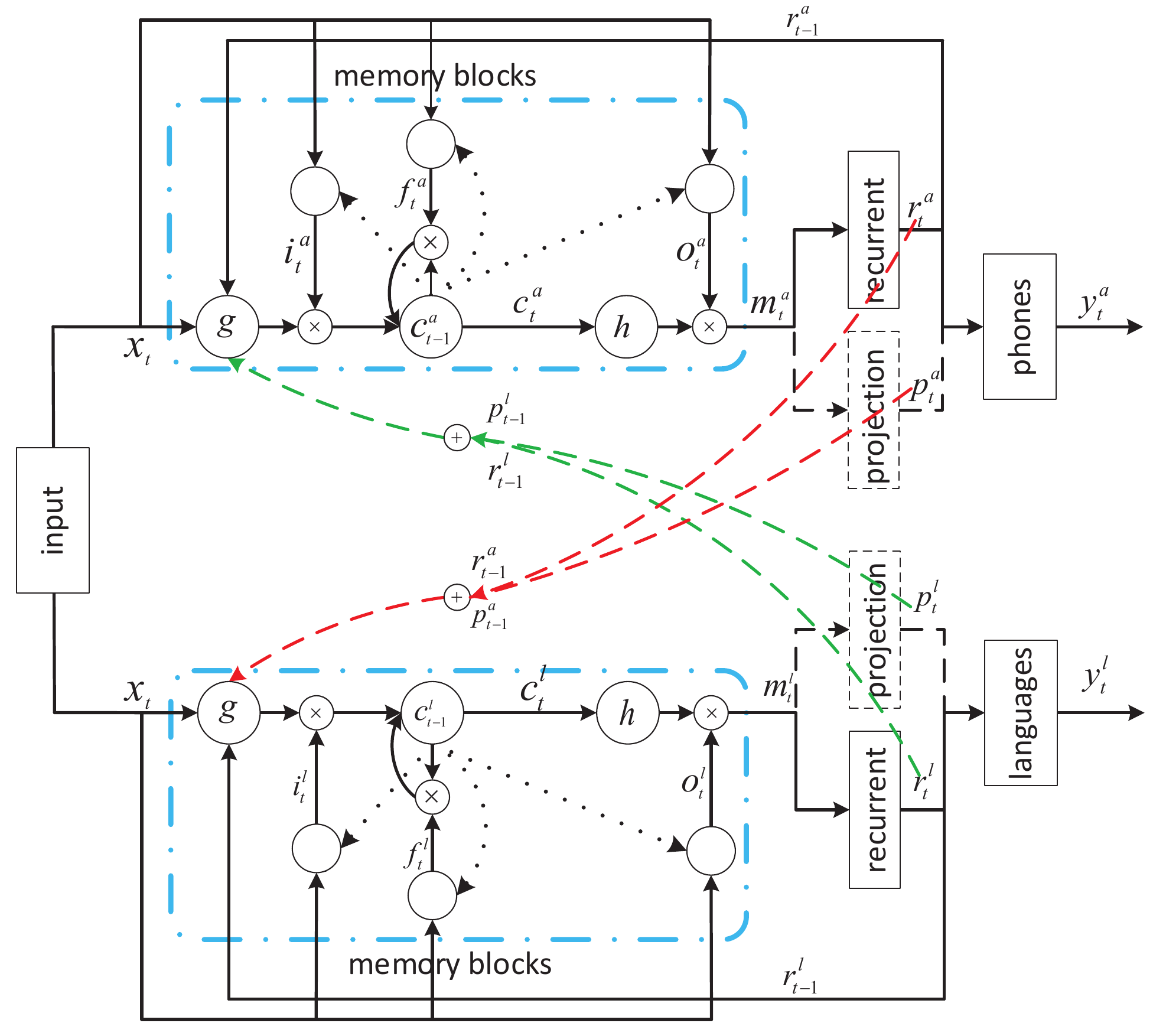}
        \caption{ Multi-task recurrent model for ASR and LR, an example.}
        \label{fig:multi}
  \end{figure}

  The basic idea of the multi-task recurrent model is to use the output of
  one task at the current frame as an auxiliary information to supervise other tasks when processing the next frame.
  As there are many alternatives that need to be carefully investigated. In this study, we use the recurrent LSTM model
  following the setting of~\cite{tang2016multi} to build the ASR
  component and the LR component, as shown in Fig.~\ref{fig:multi}. These two components are identical
  in structure and accept the same input signal. The only difference is
  that they are trained with different targets, one for phone discrimination and the other for language discrimination.
  Most importantly, there are some inter-task recurrent links that combine the two components as a single
  network, as shown by the dash lines in Fig.~\ref{fig:multi}.



  Fig.~\ref{fig:multi} is one simple example, where the recurrent information
  is extracted from both the recurrent projection $r_t$ and the nonrecurrent projection $p_t$,
  and the information is applied to the non-linear function $g(\cdot)$. We use the
  superscript $a$ and $l$ to denote the ASR and LR tasks respectively.
  The computation for ASR can be expressed as follows:

  \begin{eqnarray}
    i^a_t &=& \sigma(W^a_{ix}x_{t} + W^a_{ir}r^a_{t-1} + W^a_{ic}c^a_{t-1} + b^a_i) \nonumber\\
    f^a_t &=& \sigma(W^a_{fx}x_{t} + W^a_{fr}r^a_{t-1} + W^a_{fc}c^a_{t-1} + b^a_f) \nonumber\\
    g^a_t &=&  g(W^a_{cx}x^a_t + W^a_{cr}r^a_{t-1} + b^a_c +    \underline{ W^{al}_{cr}r^l_{t-1} + W^{al}_{cp}p^l_{t-1}  } ) \nonumber\\
    c^a_t &=& f^a_t \odot c^a_{t-1} + i^a_t \odot g^a_t \nonumber\\
    o^a_t &=& \sigma(W^a_{ox}x^a_t + W^a_{or}r^a_{t-1} + W^a_{oc}c^a_t + b^a_o) \nonumber\\
    m^a_t &=& o^a_t \odot h(c^a_t) \nonumber\\
    r^a_t &=& W^a_{rm} m^a_t \nonumber\\
    p^a_t &=& W^a_{pm} m^a_t \nonumber\\
    y^a_t &=& W^a_{yr}r^a_t + W^a_{yp}p^a_t + b^a_y \nonumber 
  \end{eqnarray}

  \noindent and the computation for LR is as follows:
  \begin{eqnarray}
    i^l_t &=& \sigma(W^l_{ix}x_{t} + W^l_{ir}r^l_{t-1} + W^l_{ic}c^l_{t-1} + b^l_i) \nonumber\\
    f^l_t &=& \sigma(W^l_{fx}x_{t} + W^l_{fr}r^l_{t-1} + W^l_{fc}c^l_{t-1} + b^l_f) \nonumber\\
    g^l_t &=& g(W^l_{cx}x^l_t + W^l_{cr}r^l_{t-1} + b^l_c +    \underline{ W^{la}_{cr}r^a_{t-1} + W^{la}_{cp}p^a_{t-1}  } ) \nonumber\\
    c^l_t &=& f^l_t \odot c^l_{t-1} + i^l_t \odot g^l_t \nonumber\\
    o^l_t &=& \sigma(W^l_{ox}x^l_t + W^l_{or}r^l_{t-1} + W^l_{oc}c^l_t + b^l_o) \nonumber\\
    m^l_t &=& o^l_t \odot h(c^l_t) \nonumber\\
    r^l_t &=& W^l_{rm} m^l_t \nonumber\\
    p^l_t &=& W^l_{pm} m^l_t \nonumber\\
    y^l_t &=& W^l_{yr}r^l_t + W^l_{yp}p^l_t + b^l_y \nonumber 
  \end{eqnarray}

  \section{Experiments}
  \label{sec:exp}

  The proposed method was tested with the Aurora4 and Thchs30 databases labelled with word transcripts.
  There are 2 language identities, one for English and the other for Chinese.
  We first present the single-task ASR baseline and then report the multi-task joint training model.
  All the experiments were conducted with the Kaldi toolkit~\cite{povey2011kaldi}.

  \subsection{Data}

  \begin{itemize}
    \item Training set: This set involves the train sets of Aurora4 and Thchs30.
     It consists of $17,137$ utterances.
     This set was used to  train the LSTM-based single-task bilingual system and the proposed
     multi-task recurrent system. The two subsets were also used to train monolingual ASR respectively.
  \end{itemize}

  \begin{itemize}
        \item Test set: This set involves `eval92' from Aurora4 for English and `test' from
        Thchs30 for Chinese. These two sets consist of $4,620$  and $2,495$ utterances
        and were used to evaluate the performance of
        ASR for English and Chinese respectively.

  \end{itemize}

  \subsection{ASR baseline}

  The ASR system was built largely following the Kaldi WSJ s5 nnet3 recipe, except that we used a single LSTM layer for simplicity. The dimension of the cell was $1,024$, and the dimensions of the recurrent and nonrecurrent projections were set to $256$.  The target delay was $5$ frames. The natural stochastic gradient descent (NSGD) algorithm was employed to train the model~\cite{povey2014parallel}. The input feature was the $40$-dimensional Fbanks, with a symmetric $2$-frame window to splice neighboring frames. The output layer consisted of $6,468$ units, equal to the total number of pdfs in the conventional GMM system that was trained to bootstrap the LSTM model.

  The baseline of monolingual ASR is presented in Table~\ref{tab:asr-base},
  where the two languages were trained and decoded separately.
  Then we present
  the baseline of bilingually-trained system in Table~\ref{tab:asr-base2}, where a unified GMM system
  was shared.
  As for the latter one, we first decoded the two languages with English and Chinese language models (LMs) respectively, denoted as `mono-LM',
  and further we merged together the two LMs with a mixture weight of $0.5$ using the tool ngram,
  so both languages can be decoded within a single unified graph built with weighted finite-state transducers, denoted as
  `bi-LM'.


      \begin{table}[th]
        \caption{\label{tab:asr-base}  Monolingual ASR baseline results.}
        \vspace{2mm}
        \centerline{
          \begin{tabular}{|c|c|c|}
            \hline
                   &Englsih & Chinese \\
            \hline
            WER\%  &12.40   & 23.45   \\
            \hline
          \end{tabular}
        }
      \end{table}

      \begin{table}[th]
        \caption{\label{tab:asr-base2}  Bilingual ASR baseline results.}
        \vspace{2mm}
        \centerline{
          \begin{tabular}{|c|c|c|}
            \hline
              Language     &Englsih & Chinese \\
              Model     &WER\%  & WER\% \\
            \hline
            Mono-LM  &16.21   & 23.82   \\
            \hline
            Bi-LM  &17.80   & 23.84   \\
            \hline
          \end{tabular}
        }
      \end{table}

  \subsection{Multi-task joint training}


  Due to the flexibility of the multi-task recurrent LSTM structure, it is not possible to
  evaluate all the configurations. We explored
  some typical ones in~\cite{tang2016multi} and report the results in Table~\ref{tab:result}.
  Note that the last configure, where
  the recurrent information is fed to all the gates and the non-linear activation
  $g(\cdot)$, is equal to
  augmenting the information to the input variable $x$.

          \begin{table}[thb!]
        \caption{\label{tab:result} Joint training results with Mono-LM.}
        \vspace{2mm}
        \centerline{
          \begin{tabular}{|cc|cccc|c|c|}
            \hline
            \multicolumn{2}{|c|}{Feedback} & \multicolumn{4}{|c|}{Feedback}              & English   & Chinese \\
            \multicolumn{2}{|c|}{Info. } & \multicolumn{4}{|c|}{Input}                   & WER\%     & EER\% \\
            \hline \hline
                         $r$ &  $p$       &  $i$   & $f$     &  $o$   &   $g$           &         &       \\
                    \hline
                    $\surd$  &            & $\surd$&         &        &                 &  16.33   &  23.96 \\
                    $\surd$  &$\surd$     & $\surd$&         &        &                 &  16.27   &  23.99 \\
                    \hline
                    $\surd$  &            &        &$\surd$  &        &                 &  16.15   &  23.97 \\
                    $\surd$  &$\surd$     &        &$\surd$  &        &                 &  16.15   &  24.01 \\
                    \hline
                    $\surd$  &            &        &         &$\surd$ &                 &  16.14   &  23.90 \\
                    $\surd$  &$\surd$     &        &         &$\surd$ &                 &  16.25   &  23.97 \\
                    \hline
                    $\surd$  &            &        &         &        & $\surd$         &  16.09   &  23.69 \\
                    $\surd$  &$\surd$     &        &         &        & $\surd$         &  16.34   &  23.81 \\
                    \hline
                    $\surd$  &            & $\surd$&$\surd$  &$\surd$ &                 &  15.65   &  23.82 \\
                    $\surd$  &$\surd$     & $\surd$&$\surd$  &$\surd$ &                 &  16.06   &  23.86 \\
                    \hline
                    $\surd$  &            & $\surd$&$\surd$  &$\surd$ &$\surd$          &  16.14   &  23.89 \\
                    $\surd$  &$\surd$     & $\surd$&$\surd$  &$\surd$ &$\surd$          &  16.32   &  24.14 \\
            \hline
          \end{tabular}
        }
      \end{table}

    \begin{table}[thb!]
        \caption{\label{tab:result2} Joint training results with Bi-LM.}
        \vspace{2mm}
        \centerline{
          \begin{tabular}{|cc|cccc|c|c|}
            \hline
            \multicolumn{2}{|c|}{Feedback} & \multicolumn{4}{|c|}{Feedback}              & English   & Chinese \\
            \multicolumn{2}{|c|}{Info. } & \multicolumn{4}{|c|}{Input}                   & WER\%     & EER\% \\
            \hline \hline
                         $r$ &  $p$       &  $i$   & $f$     &  $o$   &   $g$           &         &       \\
                    \hline
                    $\surd$  &            & $\surd$&         &        &                 &  17.81   &  24.05 \\
                    $\surd$  &$\surd$     & $\surd$&         &        &                 &  17.83   &  24.03 \\
                    \hline
                    $\surd$  &            &        &$\surd$  &        &                 &  17.62   &  24.02 \\
                    $\surd$  &$\surd$     &        &$\surd$  &        &                 &  17.71   &  23.94 \\
                    \hline
                    $\surd$  &            &        &         &$\surd$ &                 &  17.62   &  23.86 \\
                    $\surd$  &$\surd$     &        &         &$\surd$ &                 &  17.69   &  23.98 \\
                    \hline
                    $\surd$  &            &        &         &        & $\surd$         &  17.54   &  {\bf23.71} \\
                    $\surd$  &$\surd$     &        &         &        & $\surd$         &  17.80   &  23.93 \\
                    \hline
                    $\surd$  &            & $\surd$&$\surd$  &$\surd$ &                 &  {\bf17.21}   &  23.84 \\
                    $\surd$  &$\surd$     & $\surd$&$\surd$  &$\surd$ &                 &  17.53   &  23.91 \\
                    \hline
                    $\surd$  &            & $\surd$&$\surd$  &$\surd$ &$\surd$          &  17.63   &  23.93 \\
                    $\surd$  &$\surd$     & $\surd$&$\surd$  &$\surd$ &$\surd$          &  17.93   &  24.18 \\
            \hline
          \end{tabular}
        }
      \end{table}

   From the results shown in Table~\ref{tab:result} and~\ref{tab:result2} decoded with
   mono-LM and bi-LM respectively, we first observe that
   the multi-task recurrent model improves the performance of English ASR more
   than that of Chinese. We attribute this to several reasons. First, the auxiliary component
   was designed to do language recognition and expected to provide extra language information only,
   but as the English and Chinese databases are not
   from the same source, the speech signal involves too much channel information, that makes
   the effect of auxiliary language information decrease when channel classification is done at the same time.
   Moreover, the channel classification was easily achieved by the regular DNN, then the superiority
   with an additional LR component decays.
   Second, from the results in Table~\ref{tab:asr-base2}, we
   find that when using their respective LMs, English gets gains of performance, while
   that is not obvious for Chinese, even considering monolingual results in Table~\ref{tab:asr-base}.
   Results with mono-LM for Chinese in Table~\ref{tab:result2} were not far away from that of
   monolingual and bilingual baselines.
   All imply that a method for improving speech recognition
   wanting remarkable improvement for this database configuration may
   not work well.
    So it's not strange that
   the performance of Chinese could not be improved much in the enhanced model.
   Furthermore, we have done another test on part of the train set and all the
   multi-task recurrent models perform better than the baseline on both English and Chinese,
   which means the recurrent models
   overfit the train set extremely, that demonstrates the ability of the proposed model.

   We also observe that the multi-task recurrent model still has the potential to
   exceed the baseline, such as when the
   recurrent information was extracted from
   the recurrent projection and fed into the activation function, which led to a better performance for
   both English and Chinese. We suppose, with many more carefully-designed architectures, the baseline will
   be surpassed more easily.


%


  \section{Conclusions}
  \label{sec:con}

  We report a multi-task recurrent learning architecture for language-aware
  speech recognition. Primary results of the bilingual ASR experiments on
  the Aurora4/Thchs30 database demonstrated that the
  presented method can employ both commonality and diversity of different languages
  between two languages to some extent by learning ASR
  and LR models simultaneously.
  Future work involves using more ideal databases from the same source,
  developing more suitable architecture for language-aware recurrent training
  and introducing more than two languages including source-scarce ones.

\section*{Acknowledgment}
This work was supported by the National Science Foundation of China (NSFC)
Project No. 61371136, and the MESTDC PhD Foundation Project No.
20130002120011.


  \bibliographystyle{IEEEtran}
  \bibliography{joint}

\begin{thebibliography}{10}
\providecommand{\url}[1]{#1}
\csname url@samestyle\endcsname
\providecommand{\newblock}{\relax}
\providecommand{\bibinfo}[2]{#2}
\providecommand{\BIBentrySTDinterwordspacing}{\spaceskip=0pt\relax}
\providecommand{\BIBentryALTinterwordstretchfactor}{4}
\providecommand{\BIBentryALTinterwordspacing}{\spaceskip=\fontdimen2\font plus
\BIBentryALTinterwordstretchfactor\fontdimen3\font minus
  \fontdimen4\font\relax}
\providecommand{\BIBforeignlanguage}[2]{{%
\expandafter\ifx\csname l@#1\endcsname\relax
\typeout{** WARNING: IEEEtran.bst: No hyphenation pattern has been}%
\typeout{** loaded for the language `#1'. Using the pattern for}%
\typeout{** the default language instead.}%
\else
\language=\csname l@#1\endcsname
\fi
#2}}
\providecommand{\BIBdecl}{\relax}
\BIBdecl

\bibitem{hinton2012deep}
G.~Hinton, L.~Deng, D.~Yu, G.~E. Dahl, A.-r. Mohamed, N.~Jaitly, A.~Senior,
  V.~Vanhoucke, P.~Nguyen, T.~N. Sainath \emph{et~al.}, ``Deep neural networks
  for acoustic modeling in speech recognition: The shared views of four
  research groups,'' \emph{Signal Processing Magazine, IEEE}, vol.~29, no.~6,
  pp. 82--97, 2012.

\bibitem{yu2015automatic}
D.~Yu and L.~Deng, \emph{Automatic Speech Recognition - A Deep Learning
  Approach}, ser. Signals and Communication Technology.\hskip 1em plus 0.5em
  minus 0.4em\relax Springer, 2015.

\bibitem{huang2013cross}
J.-T. Huang, J.~Li, D.~Yu, L.~Deng, and Y.~Gong, ``Cross-language knowledge
  transfer using multilingual deep neural network with shared hidden layers,''
  in \emph{Proceedings of IEEE International Conference on Acoustics, Speech
  and Signal Processing (ICASSP)}.\hskip 1em plus 0.5em minus 0.4em\relax IEEE,
  2013, pp. 7304--7308.

\bibitem{Ghoshal2013multilingual}
A.~Ghoshal, P.~Swietojanski, and S.~Renals, ``Multilingual training of deep
  neural networks,'' in \emph{Proceedings of IEEE International Conference on
  Acoustics, Speech and Signal Processing (ICASSP)}.\hskip 1em plus 0.5em minus
  0.4em\relax IEEE, 2013, pp. 7319--7323.

\bibitem{Heigold2013multilingual}
G.~Heigold, V.~Vanhoucke, A.~Senior, P.~Nguyen, M.~Ranzato, M.~Devin, and
  J.~Dean, ``Multilingual acoustic models using distributed deep neural
  networks,'' in \emph{Proceedings of IEEE International Conference on
  Acoustics, Speech and Signal Processing (ICASSP)}.\hskip 1em plus 0.5em minus
  0.4em\relax IEEE, 2013, pp. 8619--8623.

\bibitem{amodei2015deep}
D.~Amodei, R.~Anubhai, E.~Battenberg, C.~Case, J.~Casper, B.~Catanzaro,
  J.~Chen, M.~Chrzanowski, A.~Coates, G.~Diamos \emph{et~al.}, ``Deep speech 2:
  End-to-end speech recognition in english and mandarin,'' \emph{arXiv preprint
  arXiv:1512.02595}, 2015.

\bibitem{tang2016multi}
Z.~Tang, L.~Li, and D.~Wang, ``Multi-task recurrent model for speech and
  speaker recognition,'' \emph{arXiv preprint arXiv:1603.09643}, 2016.

\bibitem{li2015modeling}
X.~Li and X.~Wu, ``Modeling speaker variability using long short-term memory
  networks for speech recognition,'' in \emph{Proceedings of the Annual
  Conference of International Speech Communication Association (INTERSPEECH)},
  2015.

\bibitem{Sak2014long1}
H.~Sak, A.~Senior, and F.~Beaufays, ``Long short-term memory based recurrent
  neural network architectures for large vocabulary speech recognition,''
  \emph{arXiv preprint arXiv:1402.1128}, 2014.

\bibitem{Sak2014long2}
H.~Sak, A.~W. Senior, and F.~Beaufays, ``Long short-term memory recurrent
  neural network architectures for large scale acoustic modeling,'' in
  \emph{Proceedings of the Annual Conference of International Speech
  Communication Association (INTERSPEECH)}, 2014, pp. 338--342.

\bibitem{povey2011kaldi}
D.~Povey, A.~Ghoshal, G.~Boulianne, L.~Burget, O.~Glembek, N.~Goel,
  M.~Hannemann, P.~Motlicek, Y.~Qian, and P.~Schwarz, ``The kaldi speech
  recognition toolkit,'' in \emph{Proceedings of IEEE 2011 workshop on
  automatic speech recognition and understanding}.\hskip 1em plus 0.5em minus
  0.4em\relax IEEE Signal Processing Society, 2011.

\bibitem{povey2014parallel}
D.~Povey, X.~Zhang, and S.~Khudanpur, ``Parallel training of deep neural
  networks with natural gradient and parameter averaging,'' \emph{arXiv
  preprint arXiv:1410.7455}, 2014.

\end{thebibliography}

\end{document}